\documentclass[11pt,a4paper]{article}
\pdfoutput=1 
\usepackage[hyperref]{emnlp2020}
\usepackage{times}
\usepackage{latexsym}

\usepackage{microtype}

\aclfinalcopy 

\usepackage[utf8]{inputenc}
\usepackage{todonotes}
\usepackage{amsmath}
\usepackage{mathtools}
\usepackage{hyperref}

\DeclarePairedDelimiter\floor{\lfloor}{\rfloor}
\newcommand{\nh}{\floor*{n/2}}
\newcommand{\poinc}{Poincar\'{e} }

\title{Hyperbolic Centroid Calculations for Text Classification}

\author{Aydın Gerek\\
 Huawei Turkey R\&D Center\\ Istanbul, Turkey \\
 \texttt{aydin.gerek@huawei.com} \\
 \And
 Cüneyt Ferahlar \\
 Özyeğin University\\Istanbul, Turkey\\
 \texttt{Cuneyt.ferahlar@ozyegin.edu.tr}\\
 \AND
 Bilge Şipal Sert\\
 Afiniti AI R\&D Turkey\\
 Istanbul, Turkey\\
 \texttt{bilge.sipal@afiniti.com}
 \And
 Mehmet Can Yüney \\
 Marmara University\\
 \texttt{mehmetcanyny@gmail.com}
 \And
 Onur Taşdemir \\
 Istanbul Kültür University\\
 \AND
 Zeynep Billur Kalafat \\
 Istanbul Kültür University\\
 \And
 Mert Kelkit\\
 Marmara University\\
 \texttt{mrtklkt@gmail.com}
 \And 
 Murat Can Ganiz \\
 Marmara University\\
 \texttt{murat.ganiz@}\\\texttt{marmara.edu.tr}
}

\date{}

\begin{document}
\maketitle
\begin{abstract}
  A new development in NLP is the construction of hyperbolic word embeddings.
  As opposed to their Euclidean counterparts, hyperbolic embeddings are represented not by vectors,
  but by points in hyperbolic space.
  This makes the most common basic scheme for constructing document representations,
  namely the averaging of word vectors, meaningless in the hyperbolic setting.
  We reinterpret the vector mean as the centroid of the points represented by the vectors,
  and investigate various hyperbolic centroid schemes and their effectiveness at text classification.
\end{abstract}

\section{Introduction}
Many machine learning models require that their inputs to be vector representations.
Unfortunately certain types of data are not vectors, but may be symbolic instead.
One obvious example of such a symbolic data is text.
The need for vector representations for text has led into the development of text embedding methods.
These are methods that represent some unit of text such as words, sentences, documents, characters or morphomes as vectors
(or equivalently as points in Euclidean space).
Older vector representations \cite{fftm} such as tf-idf \cite{tfidf} or PMI \cite{pmi}
have largely been replaced by the newer neural network based methods such as word2vec \cite{word2vec1,word2vec2},
GloVe \cite{glove}, doc2vec \cite{doc2vec1,doc2vec2}, and fasttext \cite{fasttext1,fasttext2}.

Graphs are another type of data for which embedding methods have been developed. 
While graph shaped data can occur in many fields such as chemistry or social network analysis,
of primary interest to us is knowledge graphs such as WordNet \cite{wordnet1,wordnet2} and word cooccurance graphs
both of which have direct applications in natural language processing.

Until recently text (and graph) embedding methods mapped words (resp. nodes) to points in Euclidean space.
However one of the oldest assumptions in machine learning, the manifold hypothesis,
states that data embedded in Euclidean space is usually not uniformly distributed in the space but forms a submanifold of that space
with an inherent non-Euclidean geometry of its own.
This leads to the concept of explicitly embedding the data in a non-Euclidean space.

One of the characteristics of Euclidean space is that it is flat, that is to say its curvature is zero everywhere.
Hyperbolic spaces form another class of constant curvature manifolds, which are not flat but have negative curvature.
The curvature of the space determines how fast the volume of a (hyper-)sphere grows as its radius $r$ increases.
In $d$-dimensional Euclidean space volume increases proportional to $r^d$.
However in hyperbolic space the increase in volume is exponential in the radius.
This makes hyperbolic space a better fit than Euclidean space for the purpose of  embedding tree-like structures and complex networks \cite{hgocn}.
It was however only recently that graph \cite{poi_emb,lor_emb} and word embedding models \cite{pglove,hskipgram} for hyperbolic spaces were developed,
along with classifiers native to hyperbolic space \cite{hnn,largeMarg}.

In this work we present simple representation methods which aggregate hyperbolic word embeddings into hyperbolic document representations.
We test the effectiveness of these representations, paired with hyperbolic classifiers,
at the task of text classification in two different languages: English and Turkish.
To the best of our knowledge this is the first study for both composing hyperbolic document representations
and hyperbolic text classification.

\section{Background and Related Work}
\subsection{Review of Hyperbolic Space}
Lines are a basic construct of Euclidean geometry.
In non-Euclidean geometries, their analogues are called geodesics.
Specifically, a geodesic is the shortest curve between two points.
Hyperbolic geometry is a non-Euclidean geometry
arising from relaxing the parallel postulate of Euclidean geometry.
In hyperbolic geometry,
given a geodesic and a point not on it,
more than one geodesic passes through the given point
without intersecting the given geodesic \cite{iversen}.
This makes hyperbolic geometry a rich type of
geometry since there are more ways to embed nonintersecting geodesics.
There are multiple models of hyperbolic space, but the two most known are the \poinc ball model, and the Lorenz model.
In this work we use the \poinc ball model.

Ungar in \cite{ungar2008} develops a gyrovector space theory of hyperbolic geometry.
In this theory while hyperbolic points are not associated with vectors,
there are analogues of vector addition and scalar multiplication,
called Mobius addition and Mobius multiplication defined as follows

\begin{equation}
x\oplus y = \frac{(1+\frac{2}{s^2} x \cdot y + \frac{1}{s^2}\Vert y \Vert^2)x + (1- \frac{1}{s^2}\Vert x\Vert^2)y}
{1+\frac{2}{s^2} x \cdot y + \frac{1}{s^4}\Vert x \Vert^2 \Vert y\Vert^2}
\end{equation}

\begin{equation}
  r\otimes\ x = s \tanh \left(r \tanh^{-1} \frac{\Vert x \Vert}{s}\right)\frac{x}{\Vert x \Vert}
\end{equation}

Using these operators a geodesic between points $a$ and $b$ in the \poinc ball model
is parametrized by $a\oplus(- a \oplus b)\otimes t$ with $0\leq t\leq 1$.
While there isn't a canonical centroid formula for $n$ points, there is a midpoint formula for $2$ points in the \poinc ball.
Picking $t=\frac{1}{2}$ gives us the midpoint $M_{ab}$ between $a$ and $b$.

\subsection{Hyperbolic Embeddings}
\poinc embeddings \cite{poi_emb} is a graph embedding technique which embeds the nodes of a complex network into a \poinc ball of arbitrary dimensionality.
One of its immediate applications was the embedding of the noun hierarchies of WordNet.
In \cite{lor_emb}, they extend this embedding technique to the Lorenz model.

However there are drawbacks of this technique for producing word embeddings,
as it relies on external resources such as WordNet whose coverage is less than that of a corpus.
As such a corpus based method was necessary, \cite{etihs} managed this by embedding the word co-occurance graph of the Text8 corpus \cite{text8}.
\cite{hskipgram} train word2vec skip-gram \cite{word2vec2} embeddings in the Lorenz model instead of explicitly embedding the co-occurance graph.
Similarly the \poinc GloVe model \cite{pglove} extends GloVe embeddings to the \poinc ball.

The \poinc GloVe model has the capability to embed words in not only hyperbolic space,
but in products of hyperbolic spaces.
A new graph embedding algorithm extending this idea was published in \cite{mcr}.
It embeds nodes in products of Eucledean, hyperbolic and spherical spaces.

Another advance in the field of hyperbolic embeddings has been the development of neural network primitives that work natively in hyperbolic space.
In \cite{hnn} the framework of gyrovector spaces \cite{ungar2008} is utilized to develop equivalents of matrix vector multiplication in the \poinc ball,
thereby constructing a feed-forward neural network layer native to the \poinc ball,
ultimately leading to the emergence of hyperbolic versions of multinomial logistic regression,
multilayer perceptron, and gated recurrent units \cite{gru}.
This is taken further in \cite{han} by introducing the attention mechanism \cite{attention} to the Lorenz model,
leading to the development of hyperbolic versions of the transformer \cite{transformer} and relation networks \cite{relnet}.

\section{Approach}
\subsection{Compositions}\label{comp}
In this section we review different ways in which a sequence of points $x_1,x_2,\dots,x_n$ in a \poinc ball can be composed into a single representation.
Many of these compositions are inspired by the averaging of Euclidean vectors, as such we will call them centroid methods.

\noindent{\bf Euclidean Mean:} As a baseline method we take the mean of the points as if they were Euclidean vectors.
\begin{equation}
EM[x_1,x_2,\dots,x_n] = \frac{x_1+x_2+\dots+x_n}{n}
\end{equation}

\noindent{\bf Naive Centroid:} Inspired by Euclidean mean, we sum the sequence using Mobius addition and then average it with Mobius scalar multiplication:
\begin{equation}
  NC[x_1,\dots,x_n]= (x_1 \oplus \dots \oplus x_n) \otimes \frac{1}{n}
\end{equation}
Unfortunately for large $n$ the addition can overflow in practice placing the sum in the boundary of the \poinc ball.
Therefore we multiply it $1-\varepsilon$ for a small $\varepsilon$ such as $10^{-5}$ if that is the case.

\begin{align}
N&C[x_1,x_2,\dots,x_n]  \nonumber \\
&=\begin{cases}
  (\bigoplus x_i) \otimes \frac{1}{n} & ; \Vert\bigoplus x_i\Vert < 1\\
  \left(\bigoplus x_i\right)(1-\varepsilon) \otimes \frac{1}{n} & ;\Vert\bigoplus x_i\Vert = 1
\end{cases}
\end{align}

\noindent{\bf Linear Forward Centroid:}
We consider a weighted midpoint where $t=\frac{m_b}{m_a+m_b}$ denoting it $M_{ab\vert m_am_b}$ with $m_x$ being the weight of point $x$.
The idea behind the linear forward centroid method is that we can assign a weight $1$ to each point $x_i$ in our sequence and repeatedly take weighted midpoint of the first two points. We can define this recursively as
\begin{align}
&LFC[(x_1,m_1),\dots,(x_{n-1},m_{n-1}),(x_n,m_n)] \nonumber\\
  &= M_{LFC[(x_1,m_1),\dots,(x_{n-1},m_{n-1})],x_n \vert m_1+\dots+m_{n-1},m_n}\\\nonumber\\
&LFC[x_1,\dots,x_n] = LFC[(x_1,1),\dots,(x_n,1)]
\end{align}

\noindent{\bf Linear Backward Centroid:} is the exactly same formula applied to the sequence in backwards order.
\begin{equation}
  LBC[x_1,x_2,\dots,x_n] = LFC[x_n,x_{n-1},\dots,x_1]
\end{equation}
Since Mobius addition is not commutative, we expect that in general
\begin{equation}
  LFC[x_1,\dots,x_n]\neq LBC[x_1,\dots,x_n]
\end{equation}

\noindent{\bf Linear Average Centroid:} is the midpoint of $LFC$ and $LBF$.
\begin{align}
  L&AC[x_1,x_2,\dots,x_n] \nonumber\\
  &= M_{LFC[x_1,x_2,\dots,x_n], LBC[x_1,x_2,\dots,x_n]}
\end{align}

\noindent{\bf Binary Tree Centroid:} Since the above methods calculate the centroid in linear time we propose using a divide and conquer method
to calculate the centroid in logarithmic time.
\begin{align}
  B&TC[(x_1,m_1),\dots, (x_n,m_n)] \nonumber\\
  &= \begin{cases}
    x_1 & \mkern-26mu;n=1\\
    M_{x_1x_2 \vert m_1m_2} & \mkern-26mu; n=2\\
    M_{BTC[(x_1,m_1),\dots,(x_{\nh},m_{\nh})]}\\
    \quad_{, BTC[(x_{\nh+1},m_{\nh+1}),\dots,(x_n,m_n)]}\\
    \quad_{\vert m_1+\dots+m_{\nh},m_{\nh+1}+\dots +m_n} & \mkern-26mu; n\geq 3
  \end{cases}
\end{align}

\subsection{Hyperbolic Classifiers}

In this section we will discuss classification methods  which we can use to classify  document  representations in hyperbolic space.
These document representations are obtained  from hyperbolic word embedding and various  document embedding models which we discuss in Section \ref{comp}.
Our tests indicate that well-known classification methods
that are based on Euclidean  optimizations and metric functions perform poorly when classifying  such vectors.
To fully observe the effect of different composition methods in creating document vectors in hyperbolic space,
we need classifiers that work in hyperbolic space. 

Our first attempt is to employ classifiers like SVM (support vector machines) with Euclidean optimizer (we use the library Libsvm  \cite{liblinear})
and a kernel function defined in hyperbolic space.    
If a kernel function or matrix satisfies Mercer conditions in hyperbolic space then  we say it is a kernel that is defined in hyperbolic space.
The easiest method to get  a kernel in hyperbolic space is to check whether any given kernel
function in Euclidean space is positive definite in hyperbolic space,
since symmetry  is straightforward \cite{Feragen}. 

In order to convert well-known kernels;
we start with the Gaussian and Laplacian kernels, for which we need a distance function in hyperbolic space.
The distance function that  we  use is  called the \poinc distance.
The \poinc distance  of two points,
say  $u$ and $v,$ that are chosen from the unit ball with the Riemann metric  is defined as 
\begin{equation}\label{Poi}
  d(u,v) =\mathrm{arccosh} \left(1+ 2\frac{\Vert u-v\Vert^2}{ (1-\Vert u\Vert^2) (1-\Vert v\Vert^2)} \right),
\end{equation}
where $\Vert.\Vert$ is the Euclidean distance.
Now with the help of the distance function we can write a general kernel function as 

\begin{equation}\label{kernelGen}
  K(u,v)= \mathrm{exp}(-\lambda (d(u,v)^q )),
\end{equation}
where  $\lambda,q>0. $
For $q=1$ Equation \ref{kernelGen} is called the geodesic Laplacian kernel and for $q=2$ it is called the geodesic Gaussian kernel.
It has been shown \cite{Feragen} that the geodesic Laplacian kernel  with the \poinc distance  satisfies Mercer conditions.
On the other hand the well-known the geodesic  Gaussian kernel fails to satisfy positive definiteness since the eigenvalues are neither  positive nor real.
Moreover,  characterizing the values of $q>2$,
for which Equation \ref{kernelGen} satisfies positive  definiteness in hyperbolic space is an open problem \cite{Feragen}.  

As a result of the discussion above our first classifier in the hyperbolic space is chosen to be the geodesic Laplacian kernel.
Theoretically, SVM with the geodesic hyperbolic kernel can do the job of finding the best decision boundary in that space,
however as the amount data increases using kernel matrices becomes computationally inefficient.  

In \cite{poi_emb} the authors tested \poinc embedding on three different datasets
(WordNet noun hierarchy, social network embedding and lexical entailment dataset).
On the WordNet dataset as evaluation metric they use average  rank (Rank) and mean average precision (MAP).
Briefly, Rank measures the rank of the distance of the observed relationship relative to ground-truth negative examples
and this value should be preferably rank 1. 
For this dataset, the results of their evaluation of the \poinc embedding show that with very small embedding dimensions the performance improvement
compared to the Euclidean embedding is massive (for dimensionality 5, Rank value of \poinc embedding is 4.9 and Euclidean embedding is 3542.3).  

This leads to the conclusion that \poinc embeddings can have reasonably low dimensionality
and still perform very well in capturing the underlying hierarchies of nouns.
Therefore  as a classifier we can use k-nearest neighbors classifier without suffering from curse of dimensionality.
This leads to the idea of applying k-nearest neighbors algorithm with the \poinc distance that is given in Equation \ref{Poi}
as our custom metric function.  

In the future we would like to implement the hyperbolic large margin classifier \cite{largeMarg},
which can be more preferable compared to k-NN and geodesic Laplacian Kernel.
The reason comes from the fact that the optimizer  Liblinear is more efficient compared to LibSVM \cite{liblinear},
and large margin classifier makes use of Liblinear.

\section{Experiments}
\subsection{Datasets}
We use five text classification datasets in two different languages: English and Turkish.
We use a variant of 20 newsgroups dataset \cite{20ng} which has 18,828 newsgroups posts, with duplicate posts from the original dataset removed.
It has 20 roughly equal sized topic classes.
The WebKB4 dataset is composed of 4 classes of web pages (with their HTML tags removed), with the class sizes being 504, 930, 1124, and 1641.
The remaining three datasets are Turkish language news classification datasets.
The 1150Haber dataset \cite{1150haber} is composed of 1150 news articles in five equal size classes.
The Hurriyet6c1k dataset \cite{hos} is composed of 6000 news articles extracted from the Turkish newspaper Hürriyet in six equal size classes.
Similarly the Milliyet4c1k dataset \cite{hos} extracted from Milliyet is composed of nine classes of 1000 articles each.
We used these relatively small datasets, due to the inefficiency of training SVM classifiers on larger sets.

\subsection{Setup}
We used the scikit-learn library in our experiments.
The composition methods explained in section \ref{comp} were all implemented in Python.
However since naive composition was quite unsuccessful (as expected) we did not place it in the tables.
The composition methods are shortened to lcf (resp. lcb, lca) for linear centroid forward (resp. backward, average);
fnw stands for the binary tree centroid, and bnw is the same applied backwards.

We use 100 dimensional embeddings, but also repeated some experiments using 30 dimensional embeddings for comparison.
Due to lack of space we have not included the latter experiments in our results tables, but do discuss them in the next section.
For English \poinc GloVe embeddings we downloaded and used pretrained embeddings \footnote{\url{https://polybox.ethz.ch/index.php/s/TzX6cXGqCX5KvAn}}.
However since there are no pretrained embeddings available for Turkish,
we trained those using the released \poinc GloVe code \footnote{\url{https://github.com/alex-tifrea/poincare_glove}} on a recent Turkish Wikipedia dump.

For the linear kernel SVM, we experimented with both the LinearSVC class, and the SVC class with kernel='linear' option,
both found in the scikit-learn library.
Due to their employment of different optimization libraries the end results can differ between these two classifiers.

\subsection{Discussion of Results}
Our experiments on Turkish datasets (1150Haber, H\"urriyet and Milliyet) indicate that
the best performing classification method for \poinc GloVe embeddings is k-NN with custom \poinc metric (see \ref{Poi}). 
For example, for 1150Haber using 30 dimensional representations we get  $91.8\%$  accuracy from k-NN $(k=5)$  with composition method lca. 
For the same dataset with dimensionality 100 we get $91.9\%$ accuracy with the same combination (k-NN+lca) and 5 neighbors.   
The same classifier with the composition method lcb (k-NN+lcb)  outperforms all the other classifiers and
the composition methods for the dataset H\"urriyet for both dimensionalities 30 and 100 with
accuracy scores $72.2\%$  and $72.93 \%$ respectively except for emean $72.4\%$ which is statistically the same $(\leq 1\%)$ result. 
For the dataset Milliyet the best combination is  (k-NN+emean)  for both dimension 30 and 100  with accuracy scores $ 71.4\%$ and $72.88\%$  respectively.  

For the Euclidean GloVe vectors  in general the best classifier is SVC (with Liblinear library optimization) and the composition method is emean. 
For example, for 1150Haber, SVC with emean has the accuracy score  $90.26\%$  with dimension 30 and $91.39\%$ using dimensionality 100. 
In this case k-NN with emean (k=11) gets statistically the same accuracy score ($91.4\%$), as well. 
For the dataset H\"urriyet  with dimensionality 100 and the data set Milliyet with dimension 30 the best classifier is SVC+emean $75.14\% $ and $63.95\%$ respectively).  
There are some exceptions to our generalization, for instance Hurriyet with 30 dimensions the best accuracy score is from SVC (linear kernel with Libsvm) and emean (accuracy $71.05\%$) and Milliyet with 100 dimensions k-NN+emean with (k=9) ( accuracy $72.3\%$).
We can deduce easily  that the best performing classifier for our Turkish datasets which is trained with \poinc GloVe vectors,  is k-NN with \poinc metric.
Furthermore, the difference between the composition methods lca, lcb and emean gives statistically same (less than $1\%$ difference) or very close accuracy scores. Hence it is not that straightforward to say a specific composition method outperforms the  other ones.  

\begin{table*}[!htpb]
\begin{center}
  \begin{tabular}{ll|l|l|l|l|l|l|l|l|l|l|}
\cline{3-12}
                         &       & \multicolumn{10}{c|}{DATASETS}                                                  \\ \hline
\multicolumn{1}{|l|}{k}  & comp  & NG E  & NG P  & W4 E  & W4 P  & 1150 E & 1150 P & H E   & H P   & M E   & M P   \\ \hline
\multicolumn{1}{|l|}{3}  & emean & 75.99 & 76.11 & 63.81 & 65.78 & 89.91  & 91.13  & 71.02 & 71.30 & 71.88 & 72.01 \\ \hline
\multicolumn{1}{|l|}{3}  & lcf   & 74.95 & 74.99 & 66.24 & 60.72 & NA     & 90.70  & NA    & 69.62 & NA    & 72.04 \\ \hline
\multicolumn{1}{|l|}{3}  & lcb   & 76.49 & 76.68 & 68.21 & 66.07 & NA     & 90.96  & NA    & 71.45 & NA    & 71.30 \\ \hline
\multicolumn{1}{|l|}{3}  & lca   & 75.92 & 75.61 & 67.26 & 63.62 & NA     & 90.52  & NA    & 70.97 & NA    & 72.12 \\ \hline
\multicolumn{1}{|l|}{3}  & fnw   & 66.21 & 66.71 & 51.60 & 47.75 & NA     & 80.35  & NA    & 61.35 & NA    & 62.73 \\ \hline
\multicolumn{1}{|l|}{3}  & bnw   & 70.37 & 70.33 & 59.09 & 54.28 & NA     & 85.39  & NA    & 67.15 & NA    & 63.20 \\ \hline
\multicolumn{1}{|l|}{5}  & emean & 76.26 & 76.97 & 64.03 & 65.70 & 91.39  & 91.39  & 70.88 & 71.88 & 72.08 & 72.70 \\ \hline
\multicolumn{1}{|l|}{5}  & lcf   & 75.78 & 75.61 & 65.85 & 60.93 & NA     & 90.87  & NA    & 70.35 & NA    & 72.60 \\ \hline
\multicolumn{1}{|l|}{5}  & lcb   & 76.85 & 77.02 & 67.88 & 65.99 & NA     & 91.65  & NA    & 72.20 & NA    & 72.03 \\ \hline
\multicolumn{1}{|l|}{5}  & lca   & 76.73 & 76.83 & 67.13 & 63.85 & NA     & 91.91  & NA    & 71.57 & NA    & 72.43 \\ \hline
\multicolumn{1}{|l|}{5}  & fnw   & 67.02 & 67.87 & 52.18 & 49.05 & NA     & 83.13  & NA    & 62.82 & NA    & 63.86 \\ \hline
\multicolumn{1}{|l|}{5}  & bnw   & 70.95 & 71.26 & 60.10 & 55.70 & NA     & 86.43  & NA    & 69.08 & NA    & 64.90 \\ \hline
\multicolumn{1}{|l|}{7}  & emean & 76.54 & 77.64 & 63.78 & 65.47 & 91.22  & 91.39  & 71.48 & 72.47 & 72.16 & 72.88 \\ \hline
\multicolumn{1}{|l|}{7}  & lcf   & 76.66 & 76.47 & 65.81 & 61.03 & NA     & 91.13  & NA    & 71.08 & NA    & 72.58 \\ \hline
\multicolumn{1}{|l|}{7}  & lcb   & 77.38 & 77.78 & 67.80 & 65.72 & NA     & 91.65  & NA    & 72.73 & NA    & 71.98 \\ \hline
\multicolumn{1}{|l|}{7}  & lca   & 77.09 & 77.38 & 66.93 & 63.59 & NA     & 91.48  & NA    & 72.18 & NA    & 72.71 \\ \hline
\multicolumn{1}{|l|}{7}  & fnw   & 67.85 & 68.06 & 52.40 & 49.22 & NA     & 82.87  & NA    & 63.65 & NA    & 64.04 \\ \hline
\multicolumn{1}{|l|}{7}  & bnw   & 71.49 & 71.64 & 60.31 & 56.01 & NA     & 87.57  & NA    & 69.03 & NA    & 65.60 \\ \hline
\multicolumn{1}{|l|}{9}  & emean & 76.57 & 77.85 & 63.58 & 65.51 & 91.48  & 91.39  & 71.80 & 72.38 & 72.30 & 72.57 \\ \hline
\multicolumn{1}{|l|}{9}  & lcf   & 77.16 & 76.47 & 65.62 & 60.61 & NA     & 90.52  & NA    & 70.98 & NA    & 72.32 \\ \hline
\multicolumn{1}{|l|}{9}  & lcb   & 77.47 & 77.99 & 67.41 & 65.45 & NA     & 91.57  & NA    & 72.92 & NA    & 71.76 \\ \hline
\multicolumn{1}{|l|}{9}  & lca   & 77.49 & 77.66 & 66.50 & 63.36 & NA     & 91.65  & NA    & 72.20 & NA    & 72.40 \\ \hline
\multicolumn{1}{|l|}{9}  & fnw   & 68.33 & 67.99 & 52.04 & 49.39 & NA     & 82.87  & NA    & 64.48 & NA    & 64.29 \\ \hline
\multicolumn{1}{|l|}{9}  & bnw   & 71.37 & 72.16 & 60.22 & 55.78 & NA     & 87.57  & NA    & 68.82 & NA    & 65.04 \\ \hline
\multicolumn{1}{|l|}{11} & emean & 76.59 & 77.90 & 63.46 & 65.19 & 91.39  & 91.04  & 71.60 & 72.32 & 72.01 & 72.51 \\ \hline
\multicolumn{1}{|l|}{11} & lcf   & 76.97 & 76.61 & 65.29 & 60.53 & NA     & 90.35  & NA    & 70.95 & NA    & 72.49 \\ \hline
\multicolumn{1}{|l|}{11} & lcb   & 77.26 & 77.85 & 67.10 & 65.23 & NA     & 91.91  & NA    & 72.93 & NA    & 71.76 \\ \hline
\multicolumn{1}{|l|}{11} & lca   & 77.35 & 77.42 & 66.41 & 63.40 & NA     & 91.13  & NA    & 72.37 & NA    & 72.28 \\ \hline
\multicolumn{1}{|l|}{11} & fnw   & 68.85 & 68.42 & 51.96 & 49.26 & NA     & 82.61  & NA    & 64.65 & NA    & 63.92 \\ \hline
\multicolumn{1}{|l|}{11} & bnw   & 71.68 & 72.49 & 59.94 & 55.96 & NA     & 87.65  & NA    & 69.50 & NA    & 65.51 \\ \hline
\end{tabular}

\end{center}
  \caption{k-NN Accuracy Scores. First few letters of each dataset are used to denote them. E stands for Euclidean GloVe, P for \poinc GloVe}
\end{table*}

Our observations on English datasets (Webkb4 and 20Newsgroup) show that  the best composition method for the document representation  that is done with  the Euclidean GloVe  is lcb. 
We should emphasize that similar to our results for Turkish datasets, the difference between methods are sometimes very insignificant. 
For example the score on the dataset Webkb4 that is trained with Euclidean GloVe is the best with SVC+lcb ($82.96\%$) whereas the score of SVC+bnw is very close ($82.83\%$).  
Moreover the score on the dataset 20Newsgroup with Euclidean GloVe is the best with SVC (linear kernel)+lcb ($69.19\%$) and SVC (linear kernel)+lca ($68.93\%$)  and also k-NN+lcb ($68.2\%$) with $k=3$ which shows that these combination of classifier and composition score basically the same.

\begin{table*}[!htpb]
\begin{center}
\label{tab:my-table}
  \begin{tabular}{ccl|l|l|l|l|l|l|}
\cline{4-9}
\multicolumn{1}{l}{}                                               & \multicolumn{1}{l}{}                           &            & \multicolumn{6}{l|}{Composition Method}       \\ \hline
\multicolumn{1}{|c|}{Embedding}                                    & \multicolumn{1}{c|}{Dataset}                   & Classifier & emean & lcf   & lcb   & lca   & fnw   & bnw   \\ \hline
\multicolumn{1}{|c|}{Euclidean GloVe}                              & \multicolumn{1}{c|}{Webkb4}                    & linearsvc  & 81.08 & 82.26 & 82.96 & 82.8  & 74.48 & 82.83 \\ \cline{3-9} 
\multicolumn{1}{|c|}{}                                             & \multicolumn{1}{c|}{}                          & svclinear  & 73.8  & 81.45 & 82.42 & 81.99 & 74.1  & 81.88 \\ \cline{3-9} 
\multicolumn{1}{|c|}{}                                             & \multicolumn{1}{c|}{}                          & rbf        & 39.05 & 65.7  & 66.2  & 66.02 & 57.67 & 57.77 \\ \hline
\multicolumn{1}{|c|}{\poinc GloVe}                  & \multicolumn{1}{c|}{Webkb4}                    & Laplacian  & 70.36 & 67.24 & 68.76 & 68.23 & 56.39 & 54.13 \\ \cline{3-9} 
\multicolumn{1}{|c|}{}                                             & \multicolumn{1}{c|}{}                          & linearsvc  & 66.49 & 64.18 & 64.65 & 64.29 & 59.2  & 64.12 \\ \cline{3-9} 
\multicolumn{1}{|c|}{}                                             & \multicolumn{1}{c|}{}                          & svclinear  & 39.05 & 39.05 & 39.05 & 39.05 & 39.05 & 39.05 \\ \cline{3-9} 
\multicolumn{1}{|c|}{}                                             & \multicolumn{1}{c|}{}                          & rbf        & 39.05 & 39.05 & 39.05 & 39.05 & 39.05 & 39.05 \\ \hline
\multicolumn{1}{|c|}{Euclidean GloVe}       & \multicolumn{1}{c|}{20NG}                      & linearsvc  & 64.59 & 63.87 & 64.66 & 64.38 & 44.09 & 59.48 \\ \cline{3-9} 
\multicolumn{1}{|c|}{}                                             & \multicolumn{1}{c|}{}                          & svclinear  & 56.98 & 68.4  & 69.19 & 68.93 & 56.48 & 65.45 \\ \cline{3-9} 
\multicolumn{1}{|c|}{}                                             & \multicolumn{1}{c|}{}                          & rbf        & 9.46  & 36.63 & 37.08 & 36.85 & 31.56 & 32.53 \\ \hline
\multicolumn{1}{|c|}{\poinc GloVe} & \multicolumn{1}{c|}{20NG}    & Laplacian  & 51.57 & 45.78 & 49.17 & 47.58 & 35.19 & 42.6  \\ \cline{3-9} 
\multicolumn{1}{|c|}{}                                             & \multicolumn{1}{c|}{}                          & linearsvc  & 59.17 & 57.11 & 59.81 & 58.68 & 44.23 & 56.45 \\ \cline{3-9} 
\multicolumn{1}{|c|}{}                                             & \multicolumn{1}{c|}{}                          & svclinear  & 8.57  & 8.58  & 8.09  & 8.35  & 6.49  & 5.31  \\ \cline{3-9} 
\multicolumn{1}{|c|}{}                                             & \multicolumn{1}{c|}{}                          & rbf        & 5.31  & 5.31  & 5.31  & 5.31  & 5.31  & 5.31  \\ \hline
\multicolumn{1}{|l|}{Euclidean GloVe}                      &\multicolumn{1}{c|} {1150Haber} & linearsvc     & 91.39  &    NA    &  NA      & NA       &      NA  &  NA      \\ \cline{3-9} 
\multicolumn{1}{|l|}{}                                 & \multicolumn{1}{l|}{}                                           & svclinear  & 89.57 &  NA      &    NA    &   NA     & NA       &    NA    \\ \cline{3-9} 
\multicolumn{1}{|l|}{}                                 & \multicolumn{1}{l|}{}                                           & rbf        & 74.7  &   NA     &      NA  &     NA   &     NA   &  NA      \\ \hline
\multicolumn{1}{|c|}{\poinc GloVe}& \multicolumn{1}{c|}{1150Haber}  & Laplacian  & 87.61 & 85.83 & 87.3  & 86.61 & 80.74 & 86.83 \\ \cline{3-9} 
\multicolumn{1}{|c|}{}                                             & \multicolumn{1}{c|}{}                          & linearsvc  & 91.26 & 90.74 & 91.04 & 91.22 & 86.65 & 90.3  \\ \cline{3-9} 
\multicolumn{1}{|c|}{}                                             & \multicolumn{1}{c|}{}                          & svclinear  & 77.13 & 74.39 & 77.17 & 76.22 & 68.09 & 75.87 \\ \cline{3-9} 
\multicolumn{1}{|c|}{}                                             & \multicolumn{1}{c|}{}                          & rbf        & 77.13 & 74.43 & 77.17 & 76.22 & 68.09 & 75.83 \\ \hline
\multicolumn{1}{|l|}{Euclidean GloVe} & \multicolumn{1}{l|} {Milliyet }                               & linearsvc  & 71.13 &   NA    &  NA     &   NA    &    NA   &  NA      \\ \cline{3-9} 
\multicolumn{1}{|l|}{}                                 & \multicolumn{1}{l|}{}                                       & svclinear  & 64.83 &    NA    &    NA    &   NA     &    NA    &   NA     \\ \cline{3-9} 
\multicolumn{1}{|l|}{}                                 & \multicolumn{1}{l|}{}                                           & rbf        & 46.94 &   NA     &  NA      &   NA     &   NA     &     NA   \\ \hline
\multicolumn{1}{|c|}{\poinc GloVe} & \multicolumn{1}{c|}{Milliyet}                                        & Laplacian  & 65.56 & 64.42 & 64.21 & 64.7  & 57.29 & 58.75 \\ \cline{3-9} 
\multicolumn{1}{|c|}{}                                             & \multicolumn{1}{c|}{}                          & linearsvc  & 64.12 & 63.43 & 63.55 & 63.68 & 57.36 & 58.73 \\ \cline{3-9} 
\multicolumn{1}{|c|}{}                                             & \multicolumn{1}{c|}{}                          & svclinear  & 47.7  & 45.76 & 47.78 & 47.12 & 30.55 & 44.52 \\ \cline{3-9} 
\multicolumn{1}{|c|}{}                                             & \multicolumn{1}{c|}{}                          & rbf        & 47.7  & 45.76 & 47.78 & 47.12 & 30.56 & 44.52 \\ \hline
\multicolumn{1}{|l|} {Euclidean GloVe }& \multicolumn{1}{l|}{H\"urriyet}                              & linearsvc  & 75.14 &    NA    &     NA   &   NA     &   NA     &  NA      \\ \cline{3-9} 
\multicolumn{1}{|l|}{}                                 & \multicolumn{1}{l|}{}                                           & svclinear  & 73.32 &   NA     &  NA      &   NA     & NA       &  NA      \\ \cline{3-9} 
\multicolumn{1}{|l|}{}                                 & \multicolumn{1}{l|}{}                                           & rbf        & 48.72 &   NA     &    NA    &  NA      &     NA   &   NA     \\ \hline
\multicolumn{1}{|c|}{\poinc GloVe} & \multicolumn{1}{c|}{H\"urriyet}& Laplacian  & 71.31 & 69.77 & 71.45 & 70.78 & 65.0  & 68.25 \\ \cline{3-9} 
\multicolumn{1}{|c|}{}                                             & \multicolumn{1}{c|}{}                          & linearsvc  & 68.25 & 66.04 & 67.72 & 67.23 & 61.78 & 65.27 \\ \cline{3-9} 
\multicolumn{1}{|c|}{}                                             & \multicolumn{1}{c|}{}                          & svclinear  & 53.27 & 48.92 & 52.82 & 51.13 & 39.74 & 43.02 \\ \cline{3-9} 
\multicolumn{1}{|c|}{}                                             & \multicolumn{1}{c|}{}                          & rbf        & 53.27 & 48.95 & 52.82 & 51.13 & 39.77 & 43.02 \\ \hline
\end{tabular}

\end{center}
 \caption{SVC- Accuracy Scores}
\end{table*}

Considering the composition methods with  \poinc GloVe, best method  seems to be emean but its scores  are statistically identical to the scores of lcb. For the \poinc GloVe vectors on the dataset 20Newsgroup the best score belongs to k-NN+lcb ($66.06\%$) with $k=3.$ If we choose emean instead of lcb the resulting score of k-NN+emean is statistically the same ($65.7\%$) with $k=3.$ 
Webkb4 is a skewed dataset but our $F_1$-micro averaged scores are the same with accuracy scores. That is why we dismiss them in our discussion. 

Observing these results, the scores of \poinc GloVe   trained Turkish datasets (1150Haber, H\"urriyet, Milliyet) (above 72\% except for 1150Haber)  seem to be better than the English datasets (Webkb4 and 20Newsgroup) (about 70\%). In essence, best score belongs to 1150Haber (91.9\%) with k-NN+lca, which is a small dataset.  We believe the morphological properties of Turkish language enable our methods with \poinc GloVe vectors to perform better than English language.

\section{Conclusion and Future Work}
We conduct the first empirical study on composition of hyperbolic document representations via various hyperbolic centroid formulations,
and tested their effectiveness at text classification in two different languages.
The hyperbolic centroid schemes we formulated were quite successful, in many cases their success surpassed that of the Euclidean baseline.
Among them the most successful was the linear backward centroid scheme,
quite often more successful than linear forward centroid, suggesting that the word order information that it inherently incorporates is useful,
and that the first few words in a document are more relevant to documentation than the last few words.
We are surprised at how successful Euclidean mean was in the hyperbolic setting,
but even more surprised that hyperbolic centroid schemes, especially lcb could be successful in the Euclidean setting.
This phenomenon bears further investigation.

In our composition schemes we only considered a uniform weighting scheme for the words.
However not all words are equally significant to the meaning of a document.
It is therefore possible to use weighting schemes such as inverse document frequency in these compositions. 

On the other hand, we want to implement hyperbolic SVC \cite{largeMarg} in order to see whether it will out perform Linear SVC for \poinc glove vectors.
Moreover, we believe that there might be a positive effect of classifiers with hyperbolic (non Euclidean) optimization. 
By the help of better performing classifiers for \poinc GloVe vectors we believe that we can construct
a better performing combination (classifier + composition method) for hyperbolic document classification.

We would also like to experiment with other composition methods such as Fréchet mean, or averaging tangent vectors to the origin attained by the logarithmic map, and then mapping their mean back to the \poinc ball with the exponential map.
Another direction of research would be to study composition methods on another model of hyperbolic geometry, such as the Lorenz model,
for which word embeddings already exist \cite{hskipgram}.


\bibliographystyle{acl_natbib}
\bibliography{hyperbolic}
\end{document}